\def\PREPRINT{1}
\documentclass{article}

\usepackage{microtype}
\usepackage{graphicx}
\usepackage{subcaption}
\usepackage{booktabs}
\usepackage{multirow}
\usepackage{xspace}
\usepackage[inline]{enumitem}
\usepackage{float}
\usepackage{hyperref}

\ifdefined\PREPRINT
  \usepackage[preprint]{icml2026}
\else
  \usepackage{icml2026}
\fi

\usepackage{amsmath}
\usepackage{amssymb}
\usepackage{mathtools}
\usepackage{amsthm}
\usepackage[capitalize,noabbrev]{cleveref}
\usepackage[textsize=tiny]{todonotes}

\newcommand{\NUM}[1]{\textcolor{blue}{[#1]}}  

\newcommand{\Tone}{TF-IDF\xspace}         
     
\newcommand{\Tthree}{e5-MLP\xspace}        
\newcommand{\Gtwo}{TF-IDF+Prop\xspace}     
\newcommand{\Gthree}{C\&S\xspace}

\newcommand{\pp}{\,pp\xspace}

\newcommand{\InflationMean}{0.30\pp\xspace}
\newcommand{\InflationMax}{0.91\pp\xspace}
\newcommand{\CoraAlignTheirs}{29.1$\pm$1.5\%\xspace}
\newcommand{\CoraAlignCorrect}{80.9$\pm$1.2\%\xspace}
\newcommand{\CoraAlignMajority}{30.4\%\xspace}
\newcommand{\CoraAlignAgreement}{18.4\%\xspace}
\newcommand{\CoraAlignN}{5\xspace}
\newcommand{\CoraAlignNodes}{2{,}692\xspace}
\newcommand{\NProtocolTasks}{24\xspace}
\newcommand{\LLaGAPubMedOwn}{95.03\%\xspace}
\newcommand{\LLaGAPubMedRGLM}{90.34\%\xspace}
\newcommand{\GraphGPTPubMedKCoT}{94.23\%\xspace}
\newcommand{\GraphGPTPubMedRGLM}{81.76\%\xspace}
\newcommand{\DiscLLaGA}{4.7\xspace}
\newcommand{\DiscGraphGPT}{12.5\xspace}
\newcommand{\MaxDisc}{12.5\xspace}
\newcommand{\KCoTCopiedCells}{32\xspace}
\newcommand{\KCoTSharedCells}{32\xspace}
\newcommand{\GraphTalksCoraSPT}{28.7--29.7\%\xspace}
\newcommand{\UniGraphGCNCora}{48.42\%\xspace}
\newcommand{\KipfGCNCora}{81.5\%\xspace}
\newcommand{\InflationPairs}{15\xspace}

\newcommand{\GThreeTasks}{24\xspace}
\newcommand{\GThreeMedianGap}{2.3\xspace}
\newcommand{\GThreeWithinThree}{15\xspace}

\newcommand{\GThreeTestable}{22\xspace}
\newcommand{\GThreeSigBelow}{12\xspace}
\newcommand{\GThreeSigAbove}{1\xspace}

\newcommand{\TThreeMedianGap}{2.8\xspace}

\icmltitlerunning{Do Reported GraphLLM Gains Survive Simple Baselines?}

\ifdefined\PREPRINT
  \AtBeginDocument{\hypersetup{pdfsubject={Preprint}}}
\fi
\begin{document}

\twocolumn[
  \icmltitle{Big Claims, Small Margins:\\ Do Reported GraphLLM Gains Survive Simple Baselines?}

  \icmlsetsymbol{equal}{*}
  \begin{icmlauthorlist}
    \icmlauthor{Michal Podstawski}{nask}
  \end{icmlauthorlist}
  \icmlaffiliation{nask}{NASK National Research Institute, Warsaw, Poland}
  \icmlcorrespondingauthor{Michal Podstawski}{michal.podstawski@nask.pl}

  \icmlkeywords{text-attributed graphs, graph neural networks, large language models, evaluation, reproducibility, benchmarks}
  \vskip 0.3in
]

\printAffiliationsAndNotice{}

\begin{abstract}
LLM-based methods for node classification on text-attributed graphs (TAGs) routinely report state-of-the-art accuracy.
We ask how these results compare with inexpensive baselines run on the same splits.
On the splits of three recent studies, we tune four baseline representatives with validation-only selection and compare them with the highest mean that each paper reports for an LLM-based method in the matched setting.
Correct~\&~Smooth on TF-IDF features trails this reference by a median of \GThreeMedianGap{}\,percentage points, and an MLP on frozen e5-large-v2 embeddings by \TThreeMedianGap{}\,points.
Correct~\&~Smooth is within 3 points of the reference or better on \GThreeWithinThree{} of \GThreeTasks{} tasks, but the reference is significantly higher on \GThreeSigBelow{} of \GThreeTestable{} testable tasks, so several reported gains remain substantial.
Reading released code next to the papers, we also document evaluation risks: selection of hyperparameters or checkpoints on the test set, labels of test-pool nodes used in training, baseline numbers reused without disclosure, and a text--label misalignment in a released Cora loader.
Reports of the same method on the same split also differ by up to \MaxDisc{} points across papers whose configurations are not documented to match.
\end{abstract}

\section{Introduction}
\label{sec:intro}

Text-attributed graphs (TAGs), such as citation networks, co-purchase graphs and hyperlinked web pages, attach natural-language text to every node.
A fast-growing line of work combines large language models (LLMs) with graph learning on TAGs: recent methods use LLMs to encode node text, predict from graph inputs, or supply supervision (\cref{sec:related}).
New methods typically report state-of-the-art node-classification accuracy.

Graph learning has seen this pattern before.
Careful re-evaluations showed that reported GNN gains often shrink under fair comparison \citep{shchur2018pitfalls,errica2020fair,platonov2023heterophily}.
They also showed that propagated features with a linear head \citep{wu2019sgc}, label propagation with a base predictor \citep{huang2021cs}, and tuned MLPs \citep{zhang2022glnn,katsman2024revisiting} are competitive on popular benchmarks.
Comparable reality checks exist in recommendation \citep{dacrema2019progress}, metric learning \citep{musgrave2020reality} and tabular learning \citep{grinsztajn2022tree}.
LLM-based methods add two complications: a much higher training and inference cost, and papers that inherit splits, features and sometimes baseline numbers from one another.

\paragraph{Question and scope.}
We ask a narrow question: \emph{when inexpensive baselines are run on the splits a paper used, with a fixed tuning budget and validation-only selection, how do they compare with the best LLM-based result that paper reports in the same setting?}
We cover three recent studies whose splits we could recover: LLM4HeG \citep{wu2025llm4heg} and KCoT on the LLaGA split \citep{xie2026kcot,chen2024llaga} ship masks, and for LLMNodeBed \citep{wu2025llmnodebed} we reconstruct the seeded splits from the released code.
Together they give \NProtocolTasks{} dataset--protocol pairs.
We also report problems in released code that we verified directly.

\paragraph{Findings.}
Several reported gains remain substantial and significant, but on many tasks the margin over a cheap baseline representative fixed by name is a few points (\cref{sec:results}).
The released code of several papers selects on the test set, trains on labels of test-pool nodes, or misaligns node texts and labels, and one paper reuses another paper's baseline numbers without disclosure; we cannot always establish whether the published tables came from this code (\cref{sec:cases}).
Finally, papers whose configurations are not documented to match report the same method on the same split up to \MaxDisc{} points apart, and KCoT reports no variance, so its margins cannot be tested.

\section{Related Work}
\label{sec:related}

\paragraph{LLMs on text-attributed graphs.}
The LLM can enhance or encode node text \citep{he2024tape}, or predict from graph tokens or serialized neighborhoods \citep{chen2024llaga,tang2024graphgpt,zhang2026rglm,xie2026kcot,wang2026purellm}.
It can also annotate, teach or judge in co-training loops \citep{xu2026gnnjudge,peng2026coteaching,wei2025pkd,wu2025llm4heg,duan2025lensgnn}, or provide the text side of graph foundation models \citep{he2025unigraph,wang2025git,zhu2025graphclip,liu2025sstag,wang2025llmbp}.

\paragraph{Benchmarks.}
CS-TAG \citep{yan2023cstag}, GLBench \citep{li2024glbench}, TSGFM \citep{chen2024tsgfm} and LLMNodeBed \citep{wu2025llmnodebed} re-run many methods under one protocol each.
LLMNodeBed finds that LLMs help most when labels are scarce.
These benchmarks do not systematically report the cheap baseline families we use: bag-of-words with a linear model, and label propagation or Correct~\&~Smooth.
Unlike them, we evaluate on each paper's own splits and compare against the numbers that paper reports.

\paragraph{Evaluation critiques and simple baselines.}
Split choice and unequal tuning distort GNN comparisons \citep{shchur2018pitfalls,errica2020fair,platonov2023heterophily}, and many benchmarks barely need structure \citep{liu2022taxonomy,bechlerspeicher2024gnnsuse,katsman2024revisiting,bechlerspeicher2025position}.
For GraphLLMs, \citet{petkar2025graphtalks} and \citet{xu2025structure} argue that current benchmarks can be solved largely from one modality, and a review proposes a reporting standard for graph learning \citep{kurashkin2026rigor}.
Strong simple baselines include propagated features with simple heads \citep{wu2019sgc,frasca2020sign,rubiomadrigal2026faf}, Correct~\&~Smooth \citep{huang2021cs}, and gradient-boosted trees on neighbor aggregates \citep{bazhenov2025graphland}.

\paragraph{Concurrent work.}
NodeGround \citep{lee2026nodeground} and \citet{platonov2026fairgfm} re-evaluate graph foundation models against tuned GNNs, without LLM-based methods and on their own splits.
A concurrent submission finds that most published gaps on standard node-classification benchmarks fall below the benchmarks' statistical resolution \citep{anon2026resolution}; it does not cover GraphLLM methods or compare numbers across papers.

\section{Evaluation Protocol}
\label{sec:protocol}

We specified the initial protocol before the pilot runs.
Later changes are logged with dates.
Among them are two reporting rules introduced after we inspected interim results: named family representatives, and a reference rule that rescales reported standard deviations.

\subsection{Splits, data and metrics}
\label{sec:protocol-splits}
For each paper we use its own train/validation/test split and its own node texts.
LLM4HeG and the LLaGA split used by KCoT ship masks; we use them as-is and check that they are disjoint and agree with the node identifiers in the instruction files.
For LLMNodeBed we re-implement the split function and seed sequence of the released code.
We have not verified that nothing else consumes the NumPy random state before the split in the original pipeline, so these splits are likely, but not proven, to be identical.
LLMNodeBed's semi-supervised WikiCS split cannot be recovered from the released code (its released masks have 20 columns), so that task is excluded.
We report the paper's own metric, accuracy, on its own test nodes.
External files are treated as untrusted: released code is read, never executed, and pickled graph objects are loaded with an allow-list of data-container types only.

\subsection{Baselines and representatives}
\label{sec:protocol-baselines}
Every baseline uses only node texts, the graph and training labels (\cref{tab:baselines}).
TF-IDF statistics and SVD projections are fitted on all nodes without labels. This is the same transductive feature access a GNN has.
Each family is reported through one representative fixed by name.
\Tone{} applies logistic regression to TF-IDF features, without the graph.
\Gtwo{} propagates TF-IDF as $[X,\hat AX,\dots,\hat A^{K}X]$ before logistic regression, with the input fixed to TF-IDF.
\Gthree{} applies Correct~\&~Smooth \citep{huang2021cs} to the \Tone{} predictions, and \Tthree{} is an MLP on frozen e5-large-v2 embeddings \citep{wang2022e5}, without the graph.
We do not report the best of our baselines per dataset as a result.

\begin{table}[t]
\centering
\caption{Baseline families and the representative reported in \cref{tab:main}.}
\label{tab:baselines}
\small
\begin{tabular}{@{}lp{0.72\columnwidth}@{}}
\toprule
Name & Representative \\
\midrule
\Tone & TF-IDF $\to$ logistic regression (no graph) \\
\Gtwo & propagated TF-IDF $\to$ logistic regression \\
\Gthree & Correct \& Smooth on the \Tone{} predictions \\
\Tthree & frozen e5-large-v2 embedding $\to$ MLP (no graph) \\
\bottomrule
\end{tabular}
\end{table}

\subsection{Tuning and selection}
\label{sec:protocol-hpo}
Each representative receives 50 trials of TPE search \citep{akiba2019optuna} over a fixed space (\cref{app:spaces}).
The trial with the best validation accuracy is selected, with ties broken by validation log-loss.
Test scores are recorded for every trial for later analysis, but selection uses validation scores only.
Each tuning run uses its own sampler and model seed.
On LLM4HeG's fixed split we match its 10 runs; on the LLaGA split, where KCoT reports no run count, we also use 10.
LLMNodeBed reports 4 runs: on its stored masks we perform 4 tuning runs, and where it re-draws splits per run we use one tuning run per re-drawn split.

\subsection{Reference and statistics}
\label{sec:protocol-stats}
The reference for each task is the highest mean that the paper reports for an LLM-based method in the matched evaluation setting: LLM4HeG Table~2, KCoT Table~2 (Single Focus, node classification) and LLMNodeBed Table~2 (semi-supervised and supervised).
It is a best-of-$N$ over methods on the paper's test set, which makes it a demanding comparison for a single fixed baseline; we do not claim that this makes our tests conservative.
LLM4HeG and LLMNodeBed compute their reported standard deviations with \texttt{np.std}, i.e.\ the population standard deviation.
We rescale these by $\sqrt{n/(n-1)}$ to sample standard deviations.
We then apply Welch's $t$-test from summary statistics, Holm-corrected over the table.
Where a paper reports no standard deviation or run count (KCoT), we give the difference only and mark significance as not assessable.
A non-significant difference is not evidence of equivalence.
Our runs and the paper's runs share a split but are not paired, so the test treats them as independent samples.
On the 12 tasks with fixed masks, the compared spread reflects training randomness only; on the 12 tasks with re-drawn splits, it also reflects variation in the partition of the observed graph.
On fixed masks the test therefore omits test-set sampling variance, which on graphs is inflated by dependence between test nodes \citep{anon2026resolution}.
These summary-statistic tests concern the evaluated runs and do not establish generalization to newly sampled graphs.

\subsection{Pipeline check}
\label{sec:protocol-v0}
With the same code we reproduced GCN \citep{kipf2017gcn} and SGC \citep{wu2019sgc} on the public Planetoid splits \citep{yang2016planetoid} (\cref{app:v0}).
This validates the training and evaluation code, not every data loader.

\section{Cheap Baselines on Published Splits}
\label{sec:results}

\begin{table*}[t]
\centering
\caption{Highest mean reported for an LLM-based method in the matched setting (reference) vs.\ our representatives, fixed by name, on the same splits (test accuracy, \%, mean{\scriptsize$\pm$}sample std).
In parentheses: ours $-$ reference.
$^{\blacktriangle}$/$^{\blacktriangledown}$: significant after Holm correction (Welch, \cref{sec:protocol-stats}).
$^\dagger$: no std or run count reported, so significance is not assessable.
Reference sources: LLM4HeG Table~2; KCoT Table~2 (Single Focus, node classification); LLMNodeBed Table~2.
Our baselines use 10 independent tuning runs on one fixed split for LLM4HeG and the LLaGA split, and 4 runs on LLMNodeBed (re-drawn splits in the supervised setting and for Books, Photo and Computer).}
\label{tab:main}
\small
\resizebox{\textwidth}{!}{\IfFileExists{generated/main_table.tex}{\begin{tabular}{@{}lrcccc@{}}
\toprule
Dataset & Reference & TF-IDF & TF-IDF+Prop & C\&S & e5-MLP \\
\midrule
\multicolumn{6}{@{}l}{\emph{LLM4HeG (NAACL'25)}} \\
Cornell & Vicuna-7B 77.6{\scriptsize$\pm$3.1} & 73.6{\scriptsize$\pm$1.4} ($-$4.0) & 62.6{\scriptsize$\pm$5.5} ($-$15.0)$^{\blacktriangledown}$ & 75.7{\scriptsize$\pm$2.5} ($-$1.9) & 80.7{\scriptsize$\pm$2.4} (+3.1) \\
Texas & Vicuna-7B 89.1{\scriptsize$\pm$3.5} & 83.2{\scriptsize$\pm$3.1} ($-$5.9)$^{\blacktriangledown}$ & 81.1{\scriptsize$\pm$6.1} ($-$8.0) & 80.9{\scriptsize$\pm$3.6} ($-$8.2)$^{\blacktriangledown}$ & 88.6{\scriptsize$\pm$2.6} ($-$0.5) \\
Wisconsin & 7B-to-560M 87.2{\scriptsize$\pm$2.6} & 81.8{\scriptsize$\pm$1.5} ($-$5.4)$^{\blacktriangledown}$ & 79.3{\scriptsize$\pm$2.5} ($-$7.9)$^{\blacktriangledown}$ & 80.4{\scriptsize$\pm$2.8} ($-$6.8)$^{\blacktriangledown}$ & 83.2{\scriptsize$\pm$3.3} ($-$4.0) \\
Actor & Vicuna-7B 76.8{\scriptsize$\pm$0.5} & 71.9{\scriptsize$\pm$0.4} ($-$4.9)$^{\blacktriangledown}$ & 73.1{\scriptsize$\pm$0.7} ($-$3.7)$^{\blacktriangledown}$ & 73.5{\scriptsize$\pm$0.8} ($-$3.3)$^{\blacktriangledown}$ & 76.0{\scriptsize$\pm$0.3} ($-$0.8)$^{\blacktriangledown}$ \\
Amazon & Bloom-1B 52.3{\scriptsize$\pm$0.6} & 55.2{\scriptsize$\pm$0.8} (+2.8)$^{\blacktriangle}$ & 47.4{\scriptsize$\pm$1.1} ($-$4.9)$^{\blacktriangledown}$ & 55.9{\scriptsize$\pm$1.9} (+3.5)$^{\blacktriangle}$ & 57.7{\scriptsize$\pm$0.9} (+5.4)$^{\blacktriangle}$ \\
\midrule
\multicolumn{6}{@{}l}{\emph{LLaGA split (KCoT, ICML'26)}} \\
Cora & KCoT 90.6$^\dagger$ & 84.6{\scriptsize$\pm$0.6} ($-$6.0) & 88.1{\scriptsize$\pm$0.6} ($-$2.6) & 89.2{\scriptsize$\pm$0.6} ($-$1.4) & 83.0{\scriptsize$\pm$0.8} ($-$7.6) \\
PubMed & KCoT 95.9$^\dagger$ & 91.0{\scriptsize$\pm$0.1} ($-$4.8) & 90.8{\scriptsize$\pm$0.4} ($-$5.1) & 91.2{\scriptsize$\pm$0.5} ($-$4.7) & 92.6{\scriptsize$\pm$0.2} ($-$3.2) \\
\midrule
\multicolumn{6}{@{}l}{\emph{LLMNodeBed semi-sup. (ICML'25)}} \\
Cora & ENGINE 84.2{\scriptsize$\pm$0.5} & 71.6{\scriptsize$\pm$0.8} ($-$12.6)$^{\blacktriangledown}$ & 81.4{\scriptsize$\pm$0.5} ($-$2.8)$^{\blacktriangledown}$ & 82.0{\scriptsize$\pm$0.5} ($-$2.3)$^{\blacktriangledown}$ & 73.7{\scriptsize$\pm$1.4} ($-$10.5)$^{\blacktriangledown}$ \\
Citeseer & ENGINE 72.1{\scriptsize$\pm$0.9} & 66.6{\scriptsize$\pm$0.5} ($-$5.5)$^{\blacktriangledown}$ & 63.3{\scriptsize$\pm$1.7} ($-$8.9)$^{\blacktriangledown}$ & 69.7{\scriptsize$\pm$0.1} ($-$2.4) & 70.9{\scriptsize$\pm$0.5} ($-$1.2) \\
PubMed & LLM\_IT 81.0{\scriptsize$\pm$0.5} & 72.7{\scriptsize$\pm$0.0} ($-$8.3)$^{\blacktriangledown}$ & 69.9{\scriptsize$\pm$0.3} ($-$11.0)$^{\blacktriangledown}$ & 76.0{\scriptsize$\pm$0.3} ($-$4.9)$^{\blacktriangledown}$ & 76.8{\scriptsize$\pm$0.9} ($-$4.2)$^{\blacktriangledown}$ \\
Instagram & GCN\_LLMEmb 67.5{\scriptsize$\pm$0.5} & 64.5{\scriptsize$\pm$0.2} ($-$3.0)$^{\blacktriangledown}$ & 65.1{\scriptsize$\pm$0.4} ($-$2.4)$^{\blacktriangledown}$ & 64.4{\scriptsize$\pm$0.3} ($-$3.1)$^{\blacktriangledown}$ & 66.5{\scriptsize$\pm$0.1} ($-$1.0) \\
Reddit & LLaGA 70.1{\scriptsize$\pm$0.4} & 57.6{\scriptsize$\pm$0.2} ($-$12.5)$^{\blacktriangledown}$ & 60.2{\scriptsize$\pm$0.1} ($-$9.9)$^{\blacktriangledown}$ & 59.2{\scriptsize$\pm$0.4} ($-$10.9)$^{\blacktriangledown}$ & 60.3{\scriptsize$\pm$0.3} ($-$9.8)$^{\blacktriangledown}$ \\
Books & TAPE 84.9{\scriptsize$\pm$0.3} & 79.7{\scriptsize$\pm$0.3} ($-$5.2)$^{\blacktriangledown}$ & 81.7{\scriptsize$\pm$0.2} ($-$3.3)$^{\blacktriangledown}$ & 82.3{\scriptsize$\pm$0.4} ($-$2.6)$^{\blacktriangledown}$ & 82.8{\scriptsize$\pm$0.2} ($-$2.1)$^{\blacktriangledown}$ \\
Photo & TAPE 86.5{\scriptsize$\pm$0.1} & 68.1{\scriptsize$\pm$0.3} ($-$18.3)$^{\blacktriangledown}$ & 84.0{\scriptsize$\pm$0.3} ($-$2.4)$^{\blacktriangledown}$ & 84.5{\scriptsize$\pm$0.3} ($-$2.0)$^{\blacktriangledown}$ & 74.6{\scriptsize$\pm$0.4} ($-$11.9)$^{\blacktriangledown}$ \\
Computer & TAPE 89.5{\scriptsize$\pm$0.0} & 65.3{\scriptsize$\pm$0.3} ($-$24.2)$^{\blacktriangledown}$ & 87.3{\scriptsize$\pm$0.4} ($-$2.3)$^{\blacktriangledown}$ & 88.0{\scriptsize$\pm$0.2} ($-$1.5)$^{\blacktriangledown}$ & 71.2{\scriptsize$\pm$0.3} ($-$18.4)$^{\blacktriangledown}$ \\
\midrule
\multicolumn{6}{@{}l}{\emph{LLMNodeBed supervised (ICML'25)}} \\
Cora & GCN\_LLMEmb 88.2{\scriptsize$\pm$2.1} & 82.0{\scriptsize$\pm$3.0} ($-$6.1) & 86.7{\scriptsize$\pm$1.4} ($-$1.4) & 86.5{\scriptsize$\pm$2.2} ($-$1.7) & 82.2{\scriptsize$\pm$2.0} ($-$5.9) \\
Citeseer & LLaGA 76.7{\scriptsize$\pm$2.0} & 74.6{\scriptsize$\pm$1.8} ($-$2.1) & 74.5{\scriptsize$\pm$0.5} ($-$2.3) & 76.5{\scriptsize$\pm$1.6} ($-$0.2) & 74.9{\scriptsize$\pm$2.2} ($-$1.8) \\
PubMed & LLM\_IT 94.2{\scriptsize$\pm$0.2} & 91.0{\scriptsize$\pm$0.5} ($-$3.2)$^{\blacktriangledown}$ & 89.9{\scriptsize$\pm$0.5} ($-$4.3)$^{\blacktriangledown}$ & 90.8{\scriptsize$\pm$0.5} ($-$3.4)$^{\blacktriangledown}$ & 92.3{\scriptsize$\pm$0.7} ($-$1.8) \\
WikiCS & TAPE 87.1{\scriptsize$\pm$0.8} & 83.9{\scriptsize$\pm$0.8} ($-$3.2)$^{\blacktriangledown}$ & 84.9{\scriptsize$\pm$0.4} ($-$2.2) & 86.1{\scriptsize$\pm$0.7} ($-$1.0) & 84.8{\scriptsize$\pm$1.6} ($-$2.4) \\
Instagram & LLaGA 69.2{\scriptsize$\pm$0.8} & 67.0{\scriptsize$\pm$0.4} ($-$2.1) & 66.9{\scriptsize$\pm$0.9} ($-$2.3) & 66.9{\scriptsize$\pm$0.5} ($-$2.3) & 68.7{\scriptsize$\pm$0.7} ($-$0.5) \\
Reddit & ENGINE 71.2{\scriptsize$\pm$0.9} & 58.9{\scriptsize$\pm$0.7} ($-$12.3)$^{\blacktriangledown}$ & 62.2{\scriptsize$\pm$0.8} ($-$9.0)$^{\blacktriangledown}$ & 60.6{\scriptsize$\pm$1.1} ($-$10.6)$^{\blacktriangledown}$ & 61.2{\scriptsize$\pm$0.4} ($-$10.0)$^{\blacktriangledown}$ \\
Books & TAPE 86.0{\scriptsize$\pm$0.7} & 81.9{\scriptsize$\pm$0.4} ($-$4.0)$^{\blacktriangledown}$ & 83.7{\scriptsize$\pm$0.5} ($-$2.2) & 84.3{\scriptsize$\pm$0.6} ($-$1.6) & 85.0{\scriptsize$\pm$0.6} ($-$1.0) \\
Photo & TAPE 87.7{\scriptsize$\pm$0.3} & 73.7{\scriptsize$\pm$0.5} ($-$14.0)$^{\blacktriangledown}$ & 86.0{\scriptsize$\pm$0.3} ($-$1.7)$^{\blacktriangledown}$ & 87.3{\scriptsize$\pm$0.6} ($-$0.4) & 76.9{\scriptsize$\pm$0.1} ($-$10.8)$^{\blacktriangledown}$ \\
Computer & TAPE 90.5{\scriptsize$\pm$0.2} & 69.2{\scriptsize$\pm$0.1} ($-$21.3)$^{\blacktriangledown}$ & 88.5{\scriptsize$\pm$0.2} ($-$2.0)$^{\blacktriangledown}$ & 90.7{\scriptsize$\pm$0.3} (+0.2) & 73.7{\scriptsize$\pm$0.6} ($-$16.8)$^{\blacktriangledown}$ \\
\bottomrule
\end{tabular}
}{\NUM{main table}}}
\end{table*}

\Cref{tab:main} lists all \NProtocolTasks{} tasks, and \cref{tab:families} summarizes each representative.
Propagation narrows the median gap: \Gtwo{} and \Gthree{} are closer to the reference than \Tone{} alone, which is far behind on Photo and Computer.
For \Gthree, the largest gaps are on Reddit in both LLMNodeBed settings, on Texas and Wisconsin, and on PubMed in all three settings.
\Tthree{}, which uses no graph, varies strongly by dataset.
For \Gthree, of the \GThreeTestable{} tasks where a test is possible, the reference is significantly higher on \GThreeSigBelow{} and significantly lower on \GThreeSigAbove{}; a non-significant difference is not a tie.
These tests do not establish generalization to newly sampled graphs (\cref{sec:protocol-stats}).
On the LLaGA split, KCoT reports no variance, so none of its margins can be tested.

\begin{table}[t]
\centering
\caption{Summary of \cref{tab:main} per representative. Gap $=$ reference $-$ ours, in points; ``Gap $\le 3$'' counts tasks within 3 points of the reference or better. $^{\blacktriangledown}$/$^{\blacktriangle}$: significantly below/above the reference, as in \cref{tab:main}; Testable: tasks where a test is possible.}
\label{tab:families}
\small
\setlength{\tabcolsep}{4pt}
\IfFileExists{generated/family_table.tex}{\begin{tabular}{@{}lrrrcr@{}}
\toprule
Name & Tasks & Median gap & Gap $\le 3$ & $^{\blacktriangledown}$\,/\,$^{\blacktriangle}$ & Testable \\
\midrule
TF-IDF & 24 & 5.5 & 3 & 17 / 1 & 22 \\
TF-IDF+Prop & 24 & 3.0 & 12 & 16 / 0 & 22 \\
C\&S & 24 & 2.3 & 15 & 12 / 1 & 22 \\
e5-MLP & 24 & 2.8 & 12 & 10 / 1 & 22 \\
\bottomrule
\end{tabular}
}{\NUM{family table}}
\end{table}

\paragraph{Do the gains survive?}
Several reported gains remain substantial under our comparison.
On many of the evaluated tasks, however, a cheap baseline representative fixed by name comes within a few points of the best LLM-based result.
The code audit in \cref{sec:cases} identifies separate evaluation risks; it does not establish that published gains arise from them.

\section{What the Released Code Shows}
\label{sec:cases}

We read the released code of several recent GraphLLM papers next to their tables.
Below are problems we verified in the source files, pinned to specific commits in \cref{app:cases}.
In some cases we cannot determine whether the published tables were produced with the released code; we say so where it applies.

\paragraph{A text--label misalignment in a released loader.}
\citet{petkar2025graphtalks} report \GraphTalksCoraSPT{} accuracy on Cora for text-only (soft-prompted LLM) models and conclude that Cora is ``structurally sufficient''.
Their released loader takes the text of node $i$ from the paper ID in row $i$ of the original \texttt{cora.content} file.
It takes the label of node $i$ from the PyG Planetoid version of Cora.
The two files order nodes differently: position by position, the two label sequences agree on only \CoraAlignAgreement{} of nodes.
Matching rows by their bag-of-words vectors yields a consistent mapping between the class labels of the two files.
We re-implemented the loading logic without executing their code, which leaves \CoraAlignNodes{} nodes after their de-duplication.
We then trained TF-IDF + logistic regression on \CoraAlignN{} of their seeded 60/20/20 splits.
With their alignment it reaches \CoraAlignTheirs{}, close to the majority-class rate of \CoraAlignMajority{}; with the correct alignment it reaches \CoraAlignCorrect{}.
In this reconstruction, text alone is highly informative on Cora, and the misalignment alone brings TF-IDF to near the majority-class rate.
We cannot establish whether this loader produced the published text-only result.

\paragraph{Selection on the test set.}
We found test-based selection at three levels.
LLM4HeG \citep{wu2025llm4heg} picks its learning-rate $\times$ weight-decay grid point by the highest mean test accuracy over ten seeds.
Sweep configurations in GIT's repository history \citep{wang2025git} maximize \texttt{final/test\_mean}.
The LLM--GNN co-teaching README \citep{peng2026coteaching} reports the ``round-best'' accuracy over training rounds that log test accuracy.
In the released baseline script of \citet{wang2026purellm}, GNN checkpoints are selected by test accuracy; we cannot establish whether the published baseline rows used this script.
We also reconstructed the LLM4HeG grid-selection procedure (same grid, seeds and epoch rule) for FAGCN, GraphSAGE and an MLP on their features.
In this reconstruction, test-based selection raises accuracy by \InflationMean{} on average and by at most \InflationMax{} over \InflationPairs{} dataset--model pairs.
We do not transfer this size to the full LLM4HeG pipeline.

\paragraph{Labels of test-pool nodes used in training.}
GNN-as-Judge \citep{xu2026gnnjudge} retrains its GNN judge on ``agreed'' nodes, where the GNN and the LLM predict the same class.
These nodes are drawn from the test-mask pool, and the retraining uses their \emph{ground-truth} labels.
The retrained GNN's predictions then become targets for LLM fine-tuning.
The same code evaluates LLM rows on 1{,}000 randomly sampled test nodes but GNN rows on the full test set.
We cannot tell whether the published tables used this code path.

\paragraph{Reused baseline numbers.}
In the node-classification part of the Single Focus block of KCoT's Table~2 \citep{xie2026kcot}, \KCoTCopiedCells{} of \KCoTSharedCells{} baseline cells are identical to LLaGA's Table~1 \citep{chen2024llaga}.
The paper states that it adopts LLaGA's settings; we found no statement that these numbers are taken from LLaGA.

\paragraph{One split, different numbers.}
On LLaGA's PubMed split, LLaGA reports \LLaGAPubMedOwn{} for itself, while RGLM \citep{zhang2026rglm} reports \LLaGAPubMedRGLM{} for LLaGA.
KCoT reports \GraphGPTPubMedKCoT{} for GraphGPT, whereas RGLM reports \GraphGPTPubMedRGLM{}.
The split is shared, but the configurations are not documented to match: RGLM's released evaluation script uses SBERT node features (\cref{app:cases}), which need not match the features behind the other numbers.
Without the full configuration, differences of \DiscLLaGA{} and \DiscGraphGPT{} points show that such numbers are not comparable across papers, not that either method is unstable.

\paragraph{Baselines far below standard results.}
UniGraph \citep{he2025unigraph} reports a supervised GCN on DeBERTa features at \UniGraphGCNCora{} on Cora with 20 labels per class. A standard GCN on bag-of-words features with the same label budget reaches \KipfGCNCora{} \citep{kipf2017gcn}; the features differ, and we have not established that the rest of the protocol is identical.

\section{Recommendations}
\label{sec:recs}

\begin{enumerate}[leftmargin=*]
  \item \textbf{Report cheap baselines on your split}, with the same tuning budget as your method: TF-IDF with a linear model, label propagation or Correct \& Smooth on top of it, and frozen embeddings from the LLM your method uses.
  \item \textbf{Select on validation only}, at every level: epochs, checkpoints, grids and rounds.
  \item \textbf{Release the split} (masks or seeded code) and the exact text version.
  \item \textbf{Report effect sizes with uncertainty}: mean, standard deviation and number of runs. A non-significant difference is not evidence of equivalence.
  \item \textbf{Mark copied numbers}, and copy only when splits, features and tuning match.
  \item \textbf{Evaluate all rows on the same test nodes.}
\end{enumerate}

\section{Limitations}
\label{sec:limitations}

We do not re-run the proposed methods; we compare against reported numbers, so our conclusions concern the published claims.
Results are conditional on the specified search spaces, the 50-trial budget and the chosen representatives.
We cover three studies whose splits we could recover. LLMNodeBed's splits are reconstructed from its released code and are likely, but not proven, to be identical, and studies without recoverable splits are not compared.
Finally, the code findings describe the released code at the pinned commits, which need not be the code that produced the published tables.

\paragraph{LLM Usage Statement.} This manuscript acknowledges the use of Claude Opus 5.5~\citep{opus55}, developed
by Anthropic, and of GPT-6.1 Sol~\citep{gpt61sol}, developed by OpenAI, to improve
language clarity, refine sentence structure, and enhance overall writing
precision. The study design, experiments, analysis, results, and conclusions remain
the sole responsibility of the authors.

\bibliography{refs}

@inproceedings{wu2025llmnodebed,
      title={When Do LLMs Help With Node Classification? A Comprehensive Analysis}, 
      author={Xixi Wu and Yifei Shen and Fangzhou Ge and Caihua Shan and Yizhu Jiao and Xiangguo Sun and Hong Cheng},
      year={2025},
      eprint={2502.00829},
      archivePrefix={arXiv},
      primaryClass={cs.LG},
      url={https://arxiv.org/abs/2502.00829},
      booktitle={ICML}
}

@inproceedings{li2024glbench,
      title={GLBench: A Comprehensive Benchmark for Graph with Large Language Models}, 
      author={Yuhan Li and Peisong Wang and Xiao Zhu and Aochuan Chen and Haiyun Jiang and Deng Cai and Victor Wai Kin Chan and Jia Li},
      year={2024},
      eprint={2407.07457},
      archivePrefix={arXiv},
      primaryClass={cs.LG},
      url={https://arxiv.org/abs/2407.07457},
      booktitle={NeurIPS Datasets and Benchmarks}
}

@inproceedings{petkar2025graphtalks,
      title={A Graph Talks, But Who's Listening? Rethinking Evaluations for Graph-Language Models}, 
      author={Soham Petkar and Hari Aakash K and Anirudh Vempati and Akshit Sinha and Ponnurangam Kumarauguru and Chirag Agarwal},
      year={2026},
      eprint={2508.20583},
      archivePrefix={arXiv},
      primaryClass={cs.CL},
      url={https://arxiv.org/abs/2508.20583},
      booktitle={Findings of ACL}
}

@inproceedings{xu2025structure,
      title={When Structure Doesn't Help: LLMs Do Not Read Text-Attributed Graphs as Effectively as We Expected}, 
      author={Haotian Xu and Yuning You and Tengfei Ma},
      year={2025},
      eprint={2511.16767},
      archivePrefix={arXiv},
      primaryClass={cs.LG},
      url={https://arxiv.org/abs/2511.16767},
      booktitle={LoG (extended abstract)}
}

@misc{katsman2024revisiting,
      title={Revisiting the Necessity of Graph Learning and Common Graph Benchmarks}, 
      author={Isay Katsman and Ethan Lou and Anna Gilbert},
      year={2024},
      eprint={2412.06173},
      archivePrefix={arXiv},
      primaryClass={cs.LG},
      url={https://arxiv.org/abs/2412.06173}, 
}

@inproceedings{bechlerspeicher2025position,
      title={Position: Graph Learning Will Lose Relevance Due To Poor Benchmarks}, 
      author={Maya Bechler-Speicher and Ben Finkelshtein and Fabrizio Frasca and Luis Müller and Jan Tönshoff and Antoine Siraudin and Viktor Zaverkin and Michael M. Bronstein and Mathias Niepert and Bryan Perozzi and Mikhail Galkin and Christopher Morris},
      year={2025},
      eprint={2502.14546},
      archivePrefix={arXiv},
      primaryClass={cs.LG},
      url={https://arxiv.org/abs/2502.14546},
      booktitle={ICML (position track)}
}

@inproceedings{bechlerspeicher2024gnnsuse,
      title={Graph Neural Networks Use Graphs When They Shouldn't}, 
      author={Maya Bechler-Speicher and Ido Amos and Ran Gilad-Bachrach and Amir Globerson},
      year={2024},
      eprint={2309.04332},
      archivePrefix={arXiv},
      primaryClass={cs.LG},
      url={https://arxiv.org/abs/2309.04332},
      booktitle={ICML}
}

@inproceedings{errica2020fair,
      title={A Fair Comparison of Graph Neural Networks for Graph Classification}, 
      author={Federico Errica and Marco Podda and Davide Bacciu and Alessio Micheli},
      year={2020},
      eprint={1912.09893},
      archivePrefix={arXiv},
      primaryClass={cs.LG},
      url={https://arxiv.org/abs/1912.09893},
      booktitle={ICLR}
}

@misc{shchur2018pitfalls,
      title={Pitfalls of Graph Neural Network Evaluation}, 
      author={Oleksandr Shchur and Maximilian Mumme and Aleksandar Bojchevski and Stephan Günnemann},
      year={2019},
      eprint={1811.05868},
      archivePrefix={arXiv},
      primaryClass={cs.LG},
      url={https://arxiv.org/abs/1811.05868}, 
}

@inproceedings{zhang2022glnn,
      title={Graph-less Neural Networks: Teaching Old MLPs New Tricks via Distillation}, 
      author={Shichang Zhang and Yozen Liu and Yizhou Sun and Neil Shah},
      year={2022},
      eprint={2110.08727},
      archivePrefix={arXiv},
      primaryClass={cs.LG},
      url={https://arxiv.org/abs/2110.08727},
      booktitle={ICLR}
}

@inproceedings{liu2022taxonomy,
      title={Taxonomy of Benchmarks in Graph Representation Learning}, 
      author={Renming Liu and Semih Cantürk and Frederik Wenkel and Sarah McGuire and Xinyi Wang and Anna Little and Leslie O'Bray and Michael Perlmutter and Bastian Rieck and Matthew Hirn and Guy Wolf and Ladislav Rampášek},
      year={2022},
      eprint={2206.07729},
      archivePrefix={arXiv},
      primaryClass={cs.LG},
      url={https://arxiv.org/abs/2206.07729},
      booktitle={LoG}
}

@inproceedings{wu2019sgc,
      title={Simplifying Graph Convolutional Networks}, 
      author={Felix Wu and Tianyi Zhang and Amauri Holanda de Souza Jr. and Christopher Fifty and Tao Yu and Kilian Q. Weinberger},
      year={2019},
      eprint={1902.07153},
      archivePrefix={arXiv},
      primaryClass={cs.LG},
      url={https://arxiv.org/abs/1902.07153},
      booktitle={ICML}
}

@inproceedings{huang2021cs,
      title={Combining Label Propagation and Simple Models Out-performs Graph Neural Networks}, 
      author={Qian Huang and Horace He and Abhay Singh and Ser-Nam Lim and Austin R. Benson},
      year={2021},
      eprint={2010.13993},
      archivePrefix={arXiv},
      primaryClass={cs.LG},
      url={https://arxiv.org/abs/2010.13993},
      booktitle={ICLR}
}

@inproceedings{kipf2017gcn,
      title={Semi-Supervised Classification with Graph Convolutional Networks}, 
      author={Thomas N. Kipf and Max Welling},
      year={2017},
      eprint={1609.02907},
      archivePrefix={arXiv},
      primaryClass={cs.LG},
      url={https://arxiv.org/abs/1609.02907},
      booktitle={ICLR}
}

@misc{frasca2020sign,
      title={SIGN: Scalable Inception Graph Neural Networks}, 
      author={Fabrizio Frasca and Emanuele Rossi and Davide Eynard and Ben Chamberlain and Michael Bronstein and Federico Monti},
      year={2020},
      eprint={2004.11198},
      archivePrefix={arXiv},
      primaryClass={cs.LG},
      url={https://arxiv.org/abs/2004.11198}, 
}

@inproceedings{he2024tape,
      title={Harnessing Explanations: LLM-to-LM Interpreter for Enhanced Text-Attributed Graph Representation Learning}, 
      author={Xiaoxin He and Xavier Bresson and Thomas Laurent and Adam Perold and Yann LeCun and Bryan Hooi},
      year={2024},
      eprint={2305.19523},
      archivePrefix={arXiv},
      primaryClass={cs.LG},
      url={https://arxiv.org/abs/2305.19523},
      booktitle={ICLR}
}

@inproceedings{chen2024llaga,
      title={LLaGA: Large Language and Graph Assistant}, 
      author={Runjin Chen and Tong Zhao and Ajay Jaiswal and Neil Shah and Zhangyang Wang},
      year={2024},
      eprint={2402.08170},
      archivePrefix={arXiv},
      primaryClass={cs.LG},
      url={https://arxiv.org/abs/2402.08170},
      booktitle={ICML}
}

@inproceedings{tang2024graphgpt,
      title={GraphGPT: Graph Instruction Tuning for Large Language Models}, 
      author={Jiabin Tang and Yuhao Yang and Wei Wei and Lei Shi and Lixin Su and Suqi Cheng and Dawei Yin and Chao Huang},
      year={2024},
      eprint={2310.13023},
      archivePrefix={arXiv},
      primaryClass={cs.CL},
      url={https://arxiv.org/abs/2310.13023},
      booktitle={SIGIR}
}

@inproceedings{xie2026kcot,
      title={Clustering as Reasoning: A $k$-Means Interpretation of Chain-of-Thought Graph Learning}, 
      author={Xuanting Xie and Zhaochen Guo and Bingheng Li and Xingtong Yu and Zhifei Liao and Zhao Kang and Yuan Fang},
      year={2026},
      eprint={2605.24867},
      archivePrefix={arXiv},
      primaryClass={cs.AI},
      url={https://arxiv.org/abs/2605.24867},
      booktitle={ICML}
}

@inproceedings{zhang2026rglm,
      title={Toward Graph-Tokenizing Large Language Models with Reconstructive Graph Instruction Tuning}, 
      author={Zhongjian Zhang and Xiao Wang and Mengmei Zhang and Jiarui Tan and Chuan Shi},
      year={2026},
      eprint={2603.01385},
      archivePrefix={arXiv},
      primaryClass={cs.CL},
      url={https://arxiv.org/abs/2603.01385},
      booktitle={WWW}
}

@inproceedings{wu2025llm4heg,
      title={Exploring the Potential of Large Language Models for Heterophilic Graphs}, 
      author={Yuxia Wu and Shujie Li and Yuan Fang and Chuan Shi},
      year={2025},
      eprint={2408.14134},
      archivePrefix={arXiv},
      primaryClass={cs.LG},
      url={https://arxiv.org/abs/2408.14134},
      booktitle={NAACL}
}

@inproceedings{duan2025lensgnn,
      title={Can Large Language Models Act as Ensembler for Multi-GNNs?}, 
      author={Hanqi Duan and Yao Cheng and Jianxiang Yu and Yao Liu and Xiang Li},
      year={2025},
      eprint={2410.16822},
      archivePrefix={arXiv},
      primaryClass={cs.AI},
      url={https://arxiv.org/abs/2410.16822},
      booktitle={EMNLP}
}

@inproceedings{wang2026purellm,
      title={Exploring Graph Learning Tasks with Pure LLMs: A Comprehensive Benchmark and Investigation}, 
      author={Yuxiang Wang and Xinnan Dai and Wenqi Fan and Yao Ma},
      year={2026},
      eprint={2502.18771},
      archivePrefix={arXiv},
      primaryClass={cs.LG},
      url={https://arxiv.org/abs/2502.18771},
      booktitle={Findings of ACL}
}

@inproceedings{xu2026gnnjudge,
      title={GNN-as-Judge: Unleashing the Power of LLMs for Graph Learning with GNN Feedback}, 
      author={Ruiyao Xu and Kaize Ding},
      year={2026},
      eprint={2604.08553},
      archivePrefix={arXiv},
      primaryClass={cs.LG},
      url={https://arxiv.org/abs/2604.08553},
      booktitle={ICLR}
}

@inproceedings{peng2026coteaching,
      title={Beyond the Golden Teacher: Enhancing Graph Learning through LLM-GNN Co-teaching}, 
      author={Zhuoyi Peng and Hanlin Gu and Lixin Fan and Yi Yang},
      year={2026},
      eprint={2606.11583},
      archivePrefix={arXiv},
      primaryClass={cs.LG},
      url={https://arxiv.org/abs/2606.11583},
      booktitle={NeurIPS}
}

@inproceedings{wei2025pkd,
      title={Preference-driven Knowledge Distillation for Few-shot Node Classification}, 
      author={Xing Wei and Chunchun Chen and Rui Fan and Xiaofeng Cao and Sourav Medya and Wei Ye},
      year={2025},
      eprint={2510.10116},
      archivePrefix={arXiv},
      primaryClass={cs.LG},
      url={https://arxiv.org/abs/2510.10116},
      booktitle={NeurIPS}
}

@inproceedings{wang2025git,
      title={Towards Graph Foundation Models: Learning Generalities Across Graphs via Task-Trees}, 
      author={Zehong Wang and Zheyuan Zhang and Tianyi Ma and Nitesh V Chawla and Chuxu Zhang and Yanfang Ye},
      year={2025},
      eprint={2412.16441},
      archivePrefix={arXiv},
      primaryClass={cs.LG},
      url={https://arxiv.org/abs/2412.16441},
      booktitle={ICML}
}

@inproceedings{wang2025llmbp,
      title={Model Generalization on Text Attribute Graphs: Principles with Large Language Models}, 
      author={Haoyu Wang and Shikun Liu and Rongzhe Wei and Pan Li},
      year={2025},
      eprint={2502.11836},
      archivePrefix={arXiv},
      primaryClass={cs.LG},
      url={https://arxiv.org/abs/2502.11836},
      booktitle={ICML}
}

@inproceedings{he2025unigraph,
      title={UniGraph: Learning a Unified Cross-Domain Foundation Model for Text-Attributed Graphs}, 
      author={Yufei He and Yuan Sui and Xiaoxin He and Bryan Hooi},
      year={2025},
      eprint={2402.13630},
      archivePrefix={arXiv},
      primaryClass={cs.LG},
      url={https://arxiv.org/abs/2402.13630},
      booktitle={KDD}
}

@inproceedings{liu2025sstag,
      title={SSTAG: Structure-Aware Self-Supervised Learning Method for Text-Attributed Graphs}, 
      author={Ruyue Liu and Rong Yin and Xiangzhen Bo and Xiaoshuai Hao and Yong Liu and Jinwen Zhong and Can Ma and Weiping Wang},
      year={2025},
      eprint={2510.01248},
      archivePrefix={arXiv},
      primaryClass={cs.CL},
      url={https://arxiv.org/abs/2510.01248},
      booktitle={NeurIPS}
}

@inproceedings{zhu2025graphclip,
      title={GraphCLIP: Enhancing Transferability in Graph Foundation Models for Text-Attributed Graphs}, 
      author={Yun Zhu and Haizhou Shi and Xiaotang Wang and Yongchao Liu and Yaoke Wang and Boci Peng and Chuntao Hong and Siliang Tang},
      year={2025},
      eprint={2410.10329},
      archivePrefix={arXiv},
      primaryClass={cs.LG},
      url={https://arxiv.org/abs/2410.10329},
      booktitle={WWW}
}

@inproceedings{dacrema2019progress,
      title={Are We Really Making Much Progress? A Worrying Analysis of Recent Neural Recommendation Approaches}, 
      author={Maurizio Ferrari Dacrema and Paolo Cremonesi and Dietmar Jannach},
      year={2019},
      eprint={1907.06902},
      archivePrefix={arXiv},
      primaryClass={cs.IR},
      doi={https://doi.org/10.1145/3298689.3347058},
      url={https://arxiv.org/abs/1907.06902},
      booktitle={RecSys}
}

@inproceedings{musgrave2020reality,
      title={A Metric Learning Reality Check}, 
      author={Kevin Musgrave and Serge Belongie and Ser-Nam Lim},
      year={2020},
      eprint={2003.08505},
      archivePrefix={arXiv},
      primaryClass={cs.CV},
      url={https://arxiv.org/abs/2003.08505},
      booktitle={ECCV}
}

@inproceedings{grinsztajn2022tree,
      title={Why do tree-based models still outperform deep learning on tabular data?}, 
      author={Léo Grinsztajn and Edouard Oyallon and Gaël Varoquaux},
      year={2022},
      eprint={2207.08815},
      archivePrefix={arXiv},
      primaryClass={cs.LG},
      url={https://arxiv.org/abs/2207.08815},
      booktitle={NeurIPS Datasets and Benchmarks}
}

@misc{bazhenov2025graphland,
      title={GraphLand: Evaluating Graph Machine Learning Models on Diverse Industrial Data}, 
      author={Gleb Bazhenov and Oleg Platonov and Liudmila Prokhorenkova},
      year={2026},
      eprint={2409.14500},
      archivePrefix={arXiv},
      primaryClass={cs.LG},
      url={https://arxiv.org/abs/2409.14500}, 
}

@inproceedings{rubiomadrigal2026faf,
      title={Fixed Aggregation Features Can Rival GNNs}, 
      author={Celia Rubio-Madrigal and Rebekka Burkholz},
      year={2026},
      eprint={2601.19449},
      archivePrefix={arXiv},
      primaryClass={cs.LG},
      url={https://arxiv.org/abs/2601.19449},
      booktitle={ICML}
}

@inproceedings{platonov2026fairgfm,
      title={A Fair Evaluation of Graph Foundation Models for Node Property Prediction}, 
      author={Oleg Platonov and Gleb Bazhenov and Dmitry Eremeev and Liudmila Prokhorenkova},
      year={2026},
      eprint={2606.24509},
      archivePrefix={arXiv},
      primaryClass={cs.LG},
      url={https://arxiv.org/abs/2606.24509},
      booktitle={ICML workshop}
}

@misc{lee2026nodeground,
      title={NodeGround: A Node Classification Benchmark in the Graph Foundation Model Era}, 
      author={Jinmo Lee and Dooho Lee and Minho Jeong and Jaemin Yoo},
      year={2026},
      eprint={2609.39673},
      archivePrefix={arXiv},
      primaryClass={cs.LG},
      url={https://arxiv.org/abs/2609.39673}, 
}

@inproceedings{platonov2023heterophily,
      title={A critical look at the evaluation of GNNs under heterophily: Are we really making progress?}, 
      author={Oleg Platonov and Denis Kuznedelev and Michael Diskin and Artem Babenko and Liudmila Prokhorenkova},
      year={2023},
      eprint={2302.11640},
      archivePrefix={arXiv},
      primaryClass={cs.LG},
      url={https://arxiv.org/abs/2302.11640},
      booktitle={ICLR}
}

@inproceedings{yang2016planetoid,
      title={Revisiting Semi-Supervised Learning with Graph Embeddings}, 
      author={Zhilin Yang and William W. Cohen and Ruslan Salakhutdinov},
      year={2016},
      eprint={1603.08861},
      archivePrefix={arXiv},
      primaryClass={cs.LG},
      url={https://arxiv.org/abs/1603.08861},
      booktitle={ICML}
}

@misc{wang2022e5,
      title={Text Embeddings by Weakly-Supervised Contrastive Pre-training}, 
      author={Liang Wang and Nan Yang and Xiaolong Huang and Binxing Jiao and Linjun Yang and Daxin Jiang and Rangan Majumder and Furu Wei},
      year={2024},
      eprint={2212.03533},
      archivePrefix={arXiv},
      primaryClass={cs.CL},
      url={https://arxiv.org/abs/2212.03533}, 
}

@inproceedings{akiba2019optuna,
      title={Optuna: A Next-generation Hyperparameter Optimization Framework}, 
      author={Takuya Akiba and Shotaro Sano and Toshihiko Yanase and Takeru Ohta and Masanori Koyama},
      year={2019},
      eprint={1907.10902},
      archivePrefix={arXiv},
      primaryClass={cs.LG},
      url={https://arxiv.org/abs/1907.10902},
      booktitle={KDD}
}

@inproceedings{chen2024tsgfm,
      title={Text-space Graph Foundation Models: Comprehensive Benchmarks and New Insights}, 
      author={Zhikai Chen and Haitao Mao and Jingzhe Liu and Yu Song and Bingheng Li and Wei Jin and Bahare Fatemi and Anton Tsitsulin and Bryan Perozzi and Hui Liu and Jiliang Tang},
      year={2024},
      eprint={2406.10727},
      archivePrefix={arXiv},
      primaryClass={cs.LG},
      url={https://arxiv.org/abs/2406.10727},
      booktitle={NeurIPS Datasets and Benchmarks}
}

@misc{anon2026resolution,
  author       = {{Anonymous}},
  title        = {The Statistical Resolution of Node-Classification Benchmarks},
  howpublished = {Submitted to ICLR 2027, \url{https://openreview.net/forum?id=kXvHmoaCGg}},
  year         = {2026}
}

@misc{opus55,
  author       = {{Anthropic}},
  title        = {{System Card}: {Claude Opus 5.5}},
  howpublished = {\url{https://www.anthropic.com/claude-opus-5-5-system-card}},
  year         = {2026}
}

@misc{gpt61sol,
  author       = {{OpenAI}},
  title        = {Addendum to {GPT-6 Astra System Card}: {GPT-6.1 Sol}},
  howpublished = {\url{https://deploymentsafety.openai.com/gpt-6-1-sol}},
  year         = {2026},
  note         = {29 September 2026}
}

@article{kurashkin2026rigor,
  title   = {Evaluation Rigor from Graph Neural Networks to Graph Foundation Models: A Systematic Review and a Four-Axis Reporting Standard},
  author  = {Kurashkin, Sergei O. and Tynchenko, Vadim S. and Borodulin, Aleksei S. and Hammoud, Ahmad and Tee, Connie},
  journal = {Machine Learning and Knowledge Extraction},
  volume  = {8}, number = {7}, pages = {194}, year = {2026},
  doi     = {10.3390/make8070194}
}

@inproceedings{yan2023cstag,
  title     = {A Comprehensive Study on Text-attributed Graphs: Benchmarking and Rethinking},
  author    = {Yan, Hao and Li, Chaozhuo and Long, Ruosong and Yan, Chao and Zhao, Jianan and Zhuang, Wenwen and Yin, Jun and Zhang, Peiyan and Han, Weihao and Sun, Hao and Deng, Weiwei and Zhang, Qi and Sun, Lichao and Xie, Xing and Wang, Senzhang},
  booktitle = {Advances in Neural Information Processing Systems 36 (Datasets and Benchmarks Track)},
  year      = {2023}
}
\bibliographystyle{icml2026}

\newpage
\appendix
\onecolumn
\section{Data, Splits and Node Order}
\label{app:data}
\begin{description}[leftmargin=0pt]
  \item[LLM4HeG.]
    \begin{itemize}
      \item Source: the authors' \texttt{dataset/*.npz} files (Git LFS) at commit \texttt{99c97e3}.
      \item Each file has a single train/val/test mask (48/32/20; Amazon 50/25/25). The ``10 splits'' in the training script are ten seeds on this split.
      \item Texts: \texttt{node\_texts}.
    \end{itemize}
  \item[LLaGA split (KCoT).]
    \begin{itemize}
      \item Source: LLaGA's \texttt{processed\_data.pt} for Cora and PubMed, from the data release linked in the authors' README, with stored random 6:2:2 masks.
      \item We assert that the stored masks equal the node ids in \texttt{sampled\_2\_10\_\{train,val,test\}.jsonl}.
      \item Cora node order equals \texttt{cora.content} (all 2{,}708 feature rows and labels coincide).
    \end{itemize}
  \item[LLMNodeBed.]
    \begin{itemize}
      \item Source: the authors' Hugging Face dataset \texttt{xxwu/LLMNodeBed}, revision \texttt{bba716f} (recorded in the download metadata).
      \item Files: \texttt{cora}, \texttt{citeseer}, \texttt{pubmed}, \texttt{wikics}, \texttt{instagram}, \texttt{reddit}, \texttt{history} (Books), \texttt{photo} and \texttt{computer}, each as \texttt{.pt}.
      \item Semi-supervised: stored masks, except Books/Photo/Computer, whose 10/10/60 splits are regenerated per run seed as in the released code.
      \item Supervised: \texttt{np.random.seed}$(s)$ followed by a shuffle and 60/20/20 cut points, for $s=0..3$. ogbn-arxiv uses the stored OGB split.
      \item Semi-supervised WikiCS is excluded because its 20-column masks cannot be indexed by the released code.
      \item The Cora/PubMed node order in this release differs from the LLaGA/TAPE order, so node ids are never mixed across releases.
    \end{itemize}
\end{description}

Seeds: 0--9 for LLM4HeG and the LLaGA split; 0--3 for LLMNodeBed.

\section{Search Spaces}
\label{app:spaces}
\begin{description}[leftmargin=0pt]
  \item[TF-IDF.]
    \begin{itemize}
      \item max\_features $\in\{2\text{k},5\text{k},10\text{k},20\text{k},50\text{k}\}$;
      \item min\_df $\in\{1,2,3,5\}$;
      \item $n$-grams $\in\{1,2\}$;
      \item sublinear tf $\in\{\text{yes},\text{no}\}$.
    \end{itemize}
  \item[Logistic regression.]
    \begin{itemize}
      \item $C\in[10^{-3},10^{3}]$ (log scale);
      \item class weights $\in\{\text{none},\text{balanced}\}$.
    \end{itemize}
  \item[Propagation (\Gtwo).]
    \begin{itemize}
      \item SVD dimension $\in\{64,128,256,512\}$;
      \item $K\in\{1,2,3\}$;
      \item normalization $\in\{\text{sym},\text{rw}\}$;
      \item self-loops $\in\{\text{yes},\text{no}\}$;
      \item each hop standardized and concatenated.
    \end{itemize}
  \item[Correct \& Smooth.]
    \begin{itemize}
      \item fixed-scale correction: training-node residuals are kept fixed while residuals propagate; training labels are clamped during smoothing;
      \item propagation over $\hat A$ with self-loops;
      \item $\alpha_1,\alpha_2\in[0,0.99]$;
      \item scale $\in[0.1,10]$ (log scale);
      \item normalization $\in\{\text{sym},\text{rw}\}$;
      \item 50 propagation steps;
      \item tuned jointly with the base model inside the same 50 trials.
    \end{itemize}
  \item[MLP (\Tthree).]
    \begin{itemize}
      \item embeddings: \texttt{intfloat/e5-large-v2} via sentence-transformers, prefix ``\texttt{query: }'', at most 512 tokens, L2-normalized;
      \item hidden size $\in\{64,128,256,512\}$;
      \item layers $\in\{1,2,3\}$;
      \item dropout $\in[0,0.8]$;
      \item learning rate $\in[10^{-3},3\cdot10^{-2}]$ (log scale);
      \item weight decay $\in[10^{-7},10^{-3}]$ (log scale);
      \item input standardization $\in\{\text{yes},\text{no}\}$;
      \item early stopping on validation accuracy (patience 100, at most 1{,}000 epochs).
    \end{itemize}
\end{description}

\section{Pipeline Validation}
\label{app:v0}
\begin{table}[H]\centering
\caption{GCN and SGC on the Planetoid public splits (accuracy, \%; GCN: mean$\pm$std over 10 seeds). Literature values: \citet{kipf2017gcn,wu2019sgc}.}
\IfFileExists{generated/v0_table.tex}{\begin{tabular}{@{}lcccc@{}}\toprule
Dataset & GCN (ours) & GCN (lit.) & SGC (ours) & SGC (lit.) \\\midrule
Cora & 81.9$\pm$0.8 & 81.5 & 81.0 & 81.0 \\
CiteSeer & 70.8$\pm$1.2 & 70.3 & 71.9 & 71.9 \\
PubMed & 79.2$\pm$0.4 & 79.0 & 78.8 & 78.9 \\
\bottomrule\end{tabular}
}{\NUM{v0}}
\end{table}

\section{Evidence for \cref{sec:cases}}
\label{app:cases}
\begin{itemize}[leftmargin=*]
  \item \textbf{Cora alignment.}
    \begin{itemize}
      \item Location: \texttt{rethinking-graph-language-evals/CLEGR} @ \texttt{72acb02}, \texttt{src/data/cora.py}.
      \item Planetoid labels are taken at l.~45 and l.~108; \texttt{cora.content} paper ids are read at l.~54--56.
      \item Each node's label is set from \texttt{common\_y[current\_idx]} at l.~169, 182 and 196, while its text is indexed by the \texttt{cora.content} row.
    \end{itemize}
  \item \textbf{LLM4HeG grid selection.}
    \begin{itemize}
      \item Location: \texttt{honey0219/LLM4HeG} @ \texttt{99c97e3}.
      \item \texttt{src/GNN/train.py} l.~104 and 127--128 print the test accuracy at the best-validation epoch.
      \item \texttt{src/GNN/parse\_results.py} l.~35--36 keeps the grid point with the highest mean of these test accuracies.
      \item l.~41 prints \texttt{np.std}, i.e.\ the population standard deviation.
    \end{itemize}
  \item \textbf{GIT sweeps.}
    \begin{itemize}
      \item Location: \texttt{Zehong-Wang/GIT} @ \texttt{207e3a4}.
      \item \texttt{sweep/citation/base.py} l.~16 and \texttt{sweep/citation/zero\_shot.py} l.~16 set \texttt{"metric": \{"goal": "maximize", "name": "final/test\_mean"\}}.
    \end{itemize}
  \item \textbf{Co-teaching.}
    \begin{itemize}
      \item Location: \texttt{realjeremybot/LLM-GNN-Coteaching} @ \texttt{98abd79}, \texttt{README.md} l.~91.
      \item The README states that round-by-round test accuracy is logged and that the round-best accuracy is reported.
    \end{itemize}
  \item \textbf{Pure-LLM benchmark baselines.}
    \begin{itemize}
      \item Location: \texttt{JensenYX/LLM-benchmarking} @ \texttt{9c3ac47}, \texttt{baseline/gcn\_sage\_gat\_node.py}.
      \item Accuracy is computed on \texttt{test\_mask} (l.~99--100).
      \item The checkpoint is saved when that accuracy improves (l.~119--123) and is reloaded at the end (l.~135).
    \end{itemize}
  \item \textbf{GNN-as-Judge.}
    \begin{itemize}
      \item Location: \texttt{rux001/GNN-as-Judge} @ \texttt{61fa4c1}.
      \item The candidate pool is \texttt{graph\_data.test\_mask} (\texttt{select\_influential\_nodes.py} l.~421).
      \item Retraining on agreed nodes uses \texttt{graph\_data.y[train\_mask]} (\texttt{node\_selection.py} l.~152--161), called at \texttt{create\_wsft.py} l.~173.
      \item The retrained predictions define the SFT/DPO data (l.~184).
      \item LLM test sampling: \texttt{create\_sft.py} l.~20 sets \texttt{--max\_test\_samples} with default 1000, used at l.~109.
    \end{itemize}
  \item \textbf{KCoT vs.\ LLaGA.} KCoT arXiv:2605.24867v1, Table~2 (Single Focus, node classification), against LLaGA arXiv:2402.08170v3, Table~1, compared cell by cell.
  \item \textbf{Same-split discrepancies and UniGraph.} LLaGA Table~1, KCoT Table~2, RGLM arXiv:2603.01385v1 Table~1, and UniGraph arXiv:2402.13630v3 Table~2. RGLM's released evaluation script sets SBERT node features (\texttt{zhongjian-zhang/RGLM} @ \texttt{e2e7442}, \texttt{scripts/eval.sh} l.~9).
\end{itemize}

\section{Baselines in Existing Benchmarks}
\label{app:benchmark-audit}
\begin{itemize}
  \item None of CS-TAG, GLBench, TSGFM or LLMNodeBed reports TF-IDF with a linear model or gradient-boosted trees without a graph, or label propagation / Correct \& Smooth, in its main results.
  \item Parameter-free propagation appears only in narrow settings: CS-TAG on ogbn-papers100M, and TSGFM as a co-trained backbone.
\end{itemize}

\end{document}